\documentclass[letterpaper]{article}
\usepackage[preprint]{aaai2027}
\usepackage[hyphens]{url}
\usepackage{natbib}
\usepackage{caption}
\usepackage{graphicx}
\usepackage{algorithm}
\usepackage{algorithmic}
\usepackage{booktabs}
\usepackage{amsmath,amssymb}
\usepackage{tabularx}

\title{DynaContext: Self-Improving Dynamic Contextualization of Optimized Prompts for Heterogeneous Parameter Extraction}

\author{
    Joe Yu\textsuperscript{\rm 1}\equalcontrib\corresponding,
    Shibin Thomas Stanley Paul\textsuperscript{\rm 2}\equalcontrib,
    Sven Mayer\textsuperscript{\rm 2,\rm 3}
}
\affiliations{
    \textsuperscript{\rm 1}Technical University of Munich, Munich, Germany\\
    \textsuperscript{\rm 2}TU Dortmund University, Dortmund, Germany\\
    \textsuperscript{\rm 3}Research Center Trustworthy Data Science and Security, Dortmund, Germany\\
    joe.yu@tum.de, shibinthomas.stanleypaul@tu-dortmund.de, sven.mayer@tu-dortmund.de
}

\begin{document}

\maketitle

\begin{abstract}

Automated prompt and skill optimization typically produces a single static instruction that is reused across inference instances until the next optimization cycle. However, this approach cannot adapt when the required context, constraints, and evidence vary from one instance to another. For instance, parameter extraction from electronic component descriptions breaks this assumption: resistors, capacitors, transistors, and connectors require different fields, unit constraints, and demonstrations, and each input provides a different evidence state. We introduce DynaContext, a framework that combines an offline-optimized extraction core, learned with GEPA or SkillOpt, with inference-time contextual adaptation and validation-gated self-improvement. DynaContext routes each item through internal, external, or fallback evidence paths and composes an item-specific prompt from the core, schema, evidence, unresolved fields, and validated demonstrations. Deterministic validation and an LLM judge gate every output, uncertain cases go to human review, and only human-verified corrections enter the demonstration memory. On a single-category benchmark, average accuracy increases from 86.6\% for the base prompt to 96.9\% for standalone SkillOpt and 98.6\% for the best DynaContext configuration. Across 850 heterogeneous gold parameter facts, average field-level F1 increases from 51.8\% for an unoptimized, demonstration-free control to 59.2\% with dynamic demonstrations alone, 66.9\% with the optimized core alone, and 71.0\% with both. Holding the model fixed, the full configuration outperforms the deployed static-prompting pipeline by 17.3 F1 points on average.
\end{abstract}

\section{Introduction}
\label{sec:introduction}

Automatic prompt and skill optimization has made LLM pipelines substantially more reliable: reflective optimizers, such as GEPA \citep{agrawal2025gepa}, evolve strong reusable instructions, program-level optimizers tune instructions and demonstrations jointly \citep{khattab2024dspy,opsahlong2024mipro}, and SkillOpt learns reusable procedural skills from execution traces \citep{yang2026skillopt}. These methods share an assumption: the optimized artifact is applied unchanged to every input until the next optimization cycle. This is effective when all inputs share a common output structure, but it breaks down for schema-heterogeneous structured prediction, where the target schema, unit constraints, relevant demonstrations, and available evidence differ from instance to instance.

Unstructured parameter extraction (UPE) for electronic components is a demanding instance of this setting. Supplier and offer data arrive as compact free-text lines such as \texttt{RES-SM,1.00K,1/10W,0.5\%,0402}, whose parameter fields are implicit. The target schema is category-conditional: a capacitor schema prioritizes capacitance, voltage, dielectric, tolerance, and package, whereas a MOSFET schema prioritizes drain-source voltage, on-resistance, current, and thermal limits. Values must separate numbers from units, every populated field must carry provenance, and unsupported fields must abstain, because the outputs feed downstream cost models and engineering decisions. Moreover, each input arrives in a different evidence state: some lines resolve to an exact internal record, some to externally verified specifications or an official datasheet, and some must be inferred from the description alone. UPE thereby extends attribute-value extraction from product text \citep{zheng2018opentag,yang2022mave}, where prompting-based methods are already competitive \citep{brinkmann2024extractgpt, brinkmann2024wdcpave} and industrial variants remain difficult \citep{mohammed2026ragsemble}.

We introduce DynaContext, a framework that separates what is optimized offline from what is composed at inference time, with self-improvement gated behind validation. GEPA and SkillOpt serve as alternative offline backends for learning a reusable extraction core. At inference, a deterministic graph with tool-using subflows \citep{yao2023react} routes each item through internal, external, or fallback evidence paths, and the prompt is composed per instance from the optimized core, the resolved schema, the gathered evidence, the unresolved fields, and retrieved validated demonstrations. The route policy is deterministic, and each route fixes its demonstration budget and fallback action, so per-instance adaptation operates inside explicit safety constraints. Unlike optimization methods that deploy a learned prompt across instances, DynaContext retains the optimized core while composing its schema, evidence, unresolved fields, and demonstrations at inference time. 

In summary, our contributions are:
\begin{itemize}
    \item We formalize UPE as schema-grounded, provenance-preserving structured prediction with mandatory abstention, and provide two benchmarks labeled by electronic cost-engineers: a heterogeneous benchmark of 133 component records spanning 13 level-3 categories 
    for a total of 850 gold parameter facts, and a 142-record single-category ceramic-capacitor set for controlled prompt and retrieval studies.

    \item We introduce DynaContext, which adapts an offline-optimized extraction core to each item through dynamic contextualization. Self-improvement is validation-gated: only human-verified corrections enter memory, and revised artifacts deploy only when they improve held-out quality without increasing false acceptances.
    \item A two-factor ablation across eight LLMs attributes the heterogeneous gain to dynamic demonstrations (+7.47 F1) and the optimized core (+15.14), which combine sub-additively to 70.98 F1, and the resulting configuration gains 17.32 F1 over the deployed pipeline while letting a small model match the strongest large-model result at a fraction of its cost (Tables~\ref{tab:main-results}--\ref{tab:cost-runtime}).
\end{itemize}

\section{Related Work}
\label{sec:related-work}

\paragraph{Attribute and parameter extraction from product text.} UPE is closest to attribute-value extraction from product profiles, which has been studied with sequence tagging \citep{zheng2018opentag}, large multi-source benchmarks \citep{yang2022mave}, and LLM-based extraction \citep{brinkmann2024extractgpt,gong2024hyperpave}. The WDC-PAVE benchmark further shows that few-shot prompting with demonstrations is a strong baseline for extraction with unit normalization \citep{brinkmann2024wdcpave}. UPE differs in three ways that drive our design: the output schema is category-conditional rather than a flat attribute set, values require value-unit separation and unit-class normalization, and every populated field must carry source provenance with abstention when evidence is missing. Universal-IE work has begun to treat schema selection itself as a learned step \citep{liang2025spt}; our catalog is instead governed and category-conditional, which is what makes abstention checkable. Closest in system spirit are retrieval-augmented structured generation for business documents \citep{cesista2024rasg} and retrieval-augmented multi-LLM ensembles for industrial part specifications \citep{mohammed2026ragsemble}, and document and datasheet pipelines that extract electrical parameters for EDA \citep{chen2024doceda,chen2025d2sflow}; DynaContext instead conditions a single optimized core per instance and gates persistence behind validated provenance rather than model consensus.

\paragraph{Automatic prompt and skill optimization.}
Reflective prompt evolution, such as GEPA \citep{agrawal2025gepa}, selects prompt candidates from a Pareto frontier using trajectory feedback, and program-level optimizers jointly tune instructions and demonstrations \citep{khattab2024dspy,opsahlong2024mipro}. Recent work extends automated prompt optimization to multimodal settings \citep{zhu2026uniapo}, online credit assignment in multi-agent systems \citep{xia2026hivemind}, adaptive defense prompts \citep{obidov2026ddpo}, and reasoning-pattern-based demonstration construction \citep{zhang2026gem}. SkillOpt \citep{yang2026skillopt} optimizes a reusable procedural skill from scored execution traces. These methods emit textual artifacts that stay fixed across inference instances until the next optimization cycle. DynaContext is complementary: it conditions the optimized artifacts on the instance schema, evidence state, and unresolved fields at inference time, and gates every artifact update behind validation.

\paragraph{Tool-using and self-improving agents.}
Interleaved reasoning and acting \citep{yao2023react} underpins tool-using agents; later work improves tool selection efficiency \citep{jia2025autotool}, execution-grounded tool retrieval \citep{wu2025gretel}, adaptive retrieval structures \citep{chen2026logicrag}, per-query weighting of sparse and dense retrieval \citep{hsu2025dat}, unified retrieval-augmented agents \citep{pham2025agentunirag}, and trustful execution over structured data \citep{zhang2025trustuqa}. Schema-grounded memory systems validate records on the write path to build durable, queryable state \citep{xmemory2026schemamemory}; DynaContext validates per-item outputs instead and treats its memory strictly as non-evidentiary demonstrations. A parallel line improves models from verification and corrective feedback, including supervised rationale verification \citep{li2025srvf}, reinforcement learning from correction signals \citep{bezalel2025tape,wu2026recode}, and self-correction distillation for structured outputs \citep{zhu2025selfcorrection}. In DynaContext, routing decisions are encoded in an optimized skill executed inside a bounded graph with fixed safety constraints, and improvements come from human-verified corrections that update the demonstration memory and the offline artifacts rather than the model weights.

\paragraph{LLM-based judging, validation, and reliability.}
LLM judges are increasingly trained and audited as evaluators \citep{huang2026thinkj,park2024offsetbias,yehudai2026clear,mohammadi2025evalmoraal,zhidkovskaya2026autopumpkin}. Guardrail frameworks enforce domain policies at system boundaries \citep{yekrangi2026gard,kumar2026infrastructuresentinel}, and reliably evaluating agentic systems in industrial and financial settings remains difficult \citep{patel2026assetopsbench,nguyen2026reliableeval}. We use the judge strictly as a triage gate over an already assembled output: automatic acceptance additionally requires deterministic validation and verified provenance, and judge verdicts are audited against human review decisions.

\section{Problem Formulation}
\label{sec:problem-formulation}

\subsection{Task Definition}
\label{subsec:task-definition}
We study source-grounded parameter extraction for heterogeneous electronic components. We define each input as a component record $x_i=(d_i,m_i,u_i,c_i,E_i)$, where $d_i$ is the raw supplier or offer description, $m_i$ is an optional manufacturer part number (MPN), $u_i$ is an optional manufacturer string, $c_i$ is an optional category hint, and $E_i$ is any current-item evidence already available to the system. The output is a closed-world structured record:
\[
\begin{aligned}
y_i=\{&\hat{c}_i,S_{\hat{c}_i},\hat{m}_i,\\
      &\{z_{if}:f\in S_{\hat{c}_i}\},M_i,q_i,\Sigma_i\}.
\end{aligned}
\]
$\hat{c}_i$ is the resolved category, $S_{\hat{c}_i}$ is the selected schema, $\hat{m}_i$ is the normalized MPN, $z_{if}$ is the value object for field $f$, $M_i$ is the set of unresolved fields, $q_i$ is the validation or review status, and $\Sigma_i$ records route and source metadata. The task is not only to fill slots: the system must select the applicable schema, determine which current-item evidence can support each field, normalize values and units, and leave unsupported schema fields null. In the heterogeneous benchmark, the active schema varies across records; in the single-component benchmark, the category and schema are fixed, isolating the extraction and context-selection part of the problem.

\subsection{Evidence Sources}
\label{subsec:evidence-sources}

Evidence is restricted to information about the current component. DynaContext distinguishes four such sources: trusted internal exact-MPN records, structured specifications returned by an external API, official-datasheet evidence acquired by the PDF Web Agent, and the raw description. These sources are considered route-conditionally, not as an exhaustive web search. An exact trusted internal match terminates retrieval and supplies the record directly. Otherwise, the system attempts MPN-grounded external retrieval through the API and, when that does not provide usable support, through official datasheets. If no reliable MPN-grounded record is available, the raw description becomes the primary evidence for fallback extraction.

Retrieved demonstrations from the validated memory are deliberately not part of $E_i$. They provide prompt context for category-specific interpretation, value--unit formatting, and null behavior, but they cannot justify a non-null parameter value for the current component. Source fusion is therefore a field-level attribution problem over current-item evidence, resolved by the route-specific priority rule of Section~\ref{sec:method}; human corrections are the only mechanism that can override an otherwise higher-ranked source.

\subsection{Output Representation}
\label{subsec:output-representation}

Each supported category is represented by a Pydantic schema that defines the allowed fields, value types, unit classes, and null behavior. A non-null parameter object has the form $z_{if}=\{\mathrm{value},\mathrm{unit},\mathrm{source}\}$: quantitative fields separate numeric value from unit, categorical fields use canonical text values, and every emitted value carries provenance. Fields outside the selected schema are rejected, even if they appear in a retrieved demonstration or neighboring category, and schema fields without sufficient current-item support remain null.

This representation is also the evaluation object. In heterogeneous multi-component experiments, each non-null gold field-value pair is treated as a parameter fact and scored with field-level precision and recall, aggregated as F1. In the ceramic-capacitor study, the same contract is evaluated within a single fixed seven-field schema, which exposes how much performance is gained when DynaContext no longer has to resolve the active component family.

\begin{figure*}[t]
    \centering
    \includegraphics[width=0.98\textwidth]{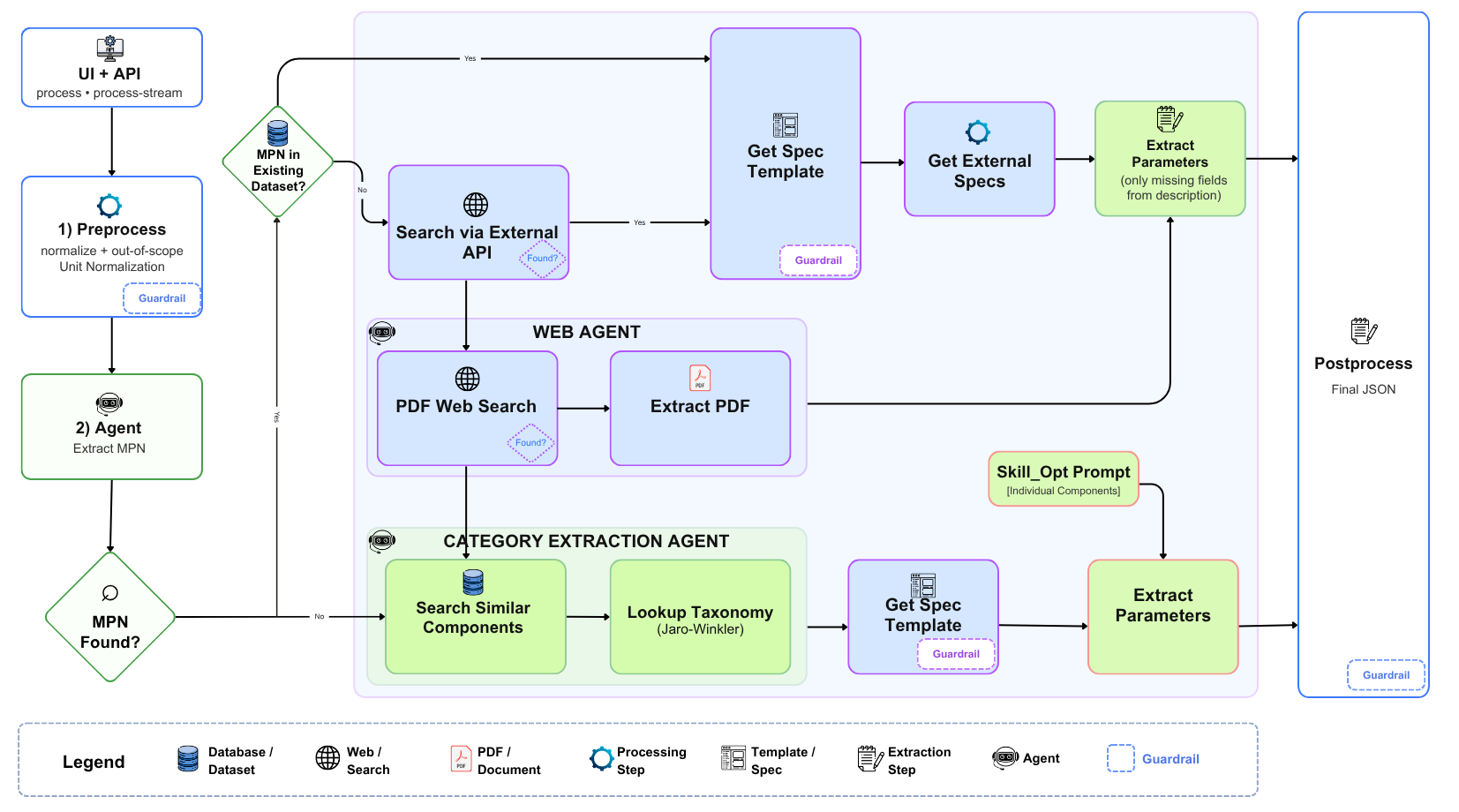}
    \caption{Route-aware DynaContext extraction workflow. Requests are preprocessed and routed by MPN availability through internal lookup, external API, and PDF retrieval, or description-based category fallback. Schema-guided extraction is followed by postprocessing into the final JSON output.}
    \label{fig:upe-flow}
\end{figure*}

\section{Method}
\label{sec:method}

\subsection{Overview and Route Selection}
DynaContext is a deterministic graph controller with localized tool-using subflows for evidence acquisition and fallback reasoning. Preprocessing normalizes the description and extracts a canonical MPN $\hat{m}_i$. Let $h_i^{I}$ indicate a trusted exact-MPN record in the internal parameter store, and $h_i^{X}$ externally verified MPN-grounded evidence from the external API or an official datasheet. The graph selects one entry route:
\[
r_i=
\begin{cases}
\mathrm{internal}, & h_i^{I}=1,\\
\mathrm{external}, & h_i^{I}=0 \land h_i^{X}=1,\\
\mathrm{fallback}, & h_i^{I}=0 \land h_i^{X}=0.
\end{cases}
\]
The checks are sequential, and Algorithm~\ref{alg:dynacontext} gives the full procedure.

\begin{algorithm}[t]
\caption{Route-aware DynaContext inference}
\label{alg:dynacontext}
\begin{algorithmic}[1]
\REQUIRE Component record $x_i$
\STATE Preprocess $d_i$, normalize units, and extract canonical MPN $\hat{m}_i$
\IF{trusted internal match exists}
  \STATE $r_i\leftarrow\mathrm{internal}$; load the complete stored record
  \STATE $M_i\leftarrow\varnothing$
\ELSIF{external API or official-datasheet evidence verifies $\hat{m}_i$}
  \STATE $r_i\leftarrow\mathrm{external}$; load verified category and parameters
  \STATE Select the category-specific specification template
  \STATE Set $M_i$ to fields missing from the external evidence
  \STATE If $M_i\neq\varnothing$, invoke the LLM to complete only those fields
\ELSE
  \STATE $r_i\leftarrow\mathrm{fallback}$; infer category and select schema
  \STATE Search similar components and match the taxonomy
  \STATE Load the specification template, the optimized core
  $A^\star_{\hat{c}_i,t}$, and demonstrations $D_i$
  \STATE $M_i\leftarrow S_{\hat{c}_i}$; extract the full schema
\ENDIF
\STATE Postprocess and validate the final JSON
\STATE Evaluate with the GEPA-optimized LLM judge; request human review when required
\STATE Persist only an accepted or human-corrected trace
\RETURN $(y_i,q_i)$
\end{algorithmic}
\end{algorithm}

On the internal route, the trusted record is reused whole: $M_i=\varnothing$ and the LLM extractor is never invoked. On the external route, the verified category selects the specification template, and retrieved values cannot be regenerated or overwritten. Neither MPN-grounded route performs similar-component search, taxonomy matching, or demonstration retrieval, which distinguishes them from the fallback route below.

\subsection{Description-Based Fallback Extraction}
\paragraph{Category and schema resolution.}
The fallback route is used when no MPN is extracted or no reliable MPN-grounded record can be established. The Category Extraction Agent searches the component store for similar items and applies Jaro--Winkler matching \citep{winkler1990jarowinkler} against the taxonomy to resolve a three-level category path. The resulting category selects the specification template and schema of Section~\ref{subsec:output-representation}. Similar items support category resolution and context selection; their parameter values are not evidence for the current component. This separates the heterogeneous problem into two stages: first, resolve the active component family, then run a category-specific extractor over only the schema fields that can be valid for that family.

\paragraph{Demonstration retrieval and fallback extraction.}
Only on this route does DynaContext query the validated demonstration memory $\mathcal{M}_t$. It retrieves the top category-matched demonstrations using embedding similarity, BM25 overlap \citep{robertson2009bm25}, or a deduplicated hybrid ranking, subject to the non-evidentiary rule of Section~\ref{subsec:evidence-sources}. The selected specification template, the optimized extraction core (the SkillOpt prompt in the deployed configuration of Figure~\ref{fig:upe-flow}), and the retrieved demonstrations condition full extraction over the selected schema.

\subsection{Schema-Constrained Extraction and Output Assembly}
\paragraph{Branch-aware prompt composition.}
Let $A_{\hat{c}_i,t}^\star$ denote the optimized extraction core for the resolved category at iteration $t$, and let $R_k(x_i,\mathcal{M}_t)$ return up to $k$ validated demonstrations from memory. DynaContext does not deploy the optimized core as a static prompt. Whenever LLM extraction is required, it composes the final prompt at inference time:
\[
\begin{aligned}
P_t(x_i)={}&A_{\hat{c}_i,t}^\star \oplus S_{\hat{c}_i}
            \oplus M_i \\
            &\oplus \phi(E_i)\oplus d_i
            \oplus R_k(x_i,\mathcal{M}_t).
\end{aligned}
\]
For the internal route, $M_i=\varnothing$ and no extraction prompt is instantiated. For the external route, $M_i$ contains only fields unresolved by verified evidence and $k=0$, so the LLM can only complete missing fields from the current description. For fallback, $M_i=S_{\hat{c}_i}$ and $k>0$, so the prompt includes category-matched demonstrations for full schema extraction. Thus, DynaContext keeps the optimized core stable while dynamically changing the schema, missing-field set, evidence summary, and demonstrations for each item.

\paragraph{Structured output.}
The model emits the record of Section~\ref{subsec:output-representation}, and a deterministic post-processing step then applies the final guardrails, standardizes units, and serializes the result to JSON.

\paragraph{Source fusion.}
Fusion is route-specific rather than an instruction to query every source. On the internal route, the complete trusted record is retained as-is. On the external route, external API values precede official datasheet values, which precede LLM completion of still-missing fields. The fallback route uses the schema-constrained description extraction. For each field, the highest-priority usable candidate is retained; a candidate must be schema-valid, unit-compatible, and attributable to the current item. Lower-priority candidates fill only null fields. Incompatible non-null values are logged as conflicts and sent to review.

\subsection{Validation, Human Review, and Self-Improvement}
\begin{figure}[t]
    \centering
    \includegraphics[width=0.98\columnwidth]{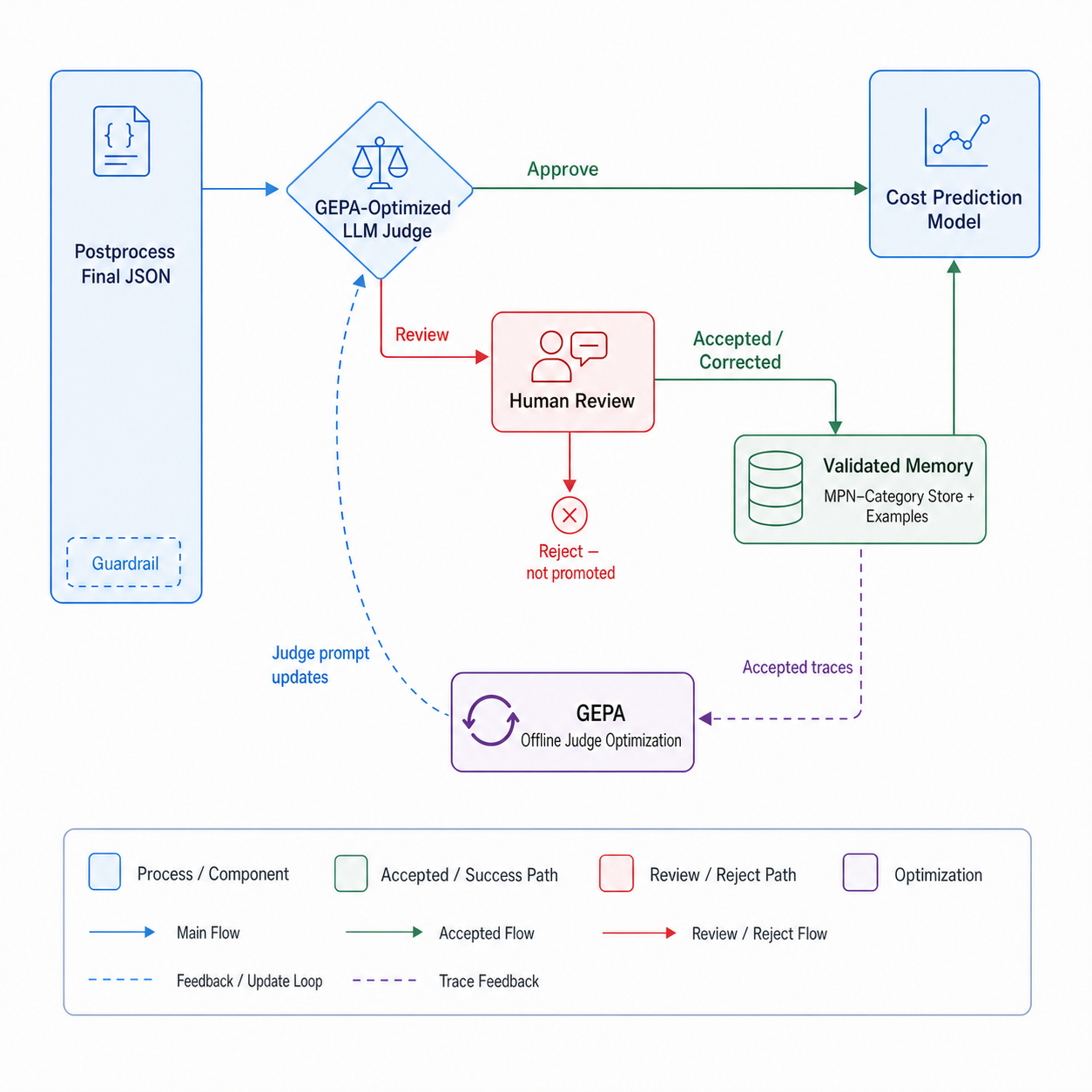}
    \caption{Validation, human-review, and self-improvement loop. Accepted
    corrections update validated memory and feed the offline GEPA
    judge-optimization cycle; rejected cases are not promoted.}
    \label{fig:validation-feedback}
\end{figure}

Figure~\ref{fig:validation-feedback} summarizes the validation, review, and offline self-improvement loop.

\paragraph{Deterministic validation.}
Guardrails are applied during pre-processing, specification-template selection, and final post-processing. They reject fields outside the schema, malformed value--unit pairs, incompatible units, invalid MPN normalization, category--schema mismatches, unsupported non-null values, or missing provenance. Generated candidates are checked again after output assembly.

\paragraph{GEPA-optimized judge and human review.}
The LLM judge uses a prompt optimized offline with GEPA. It receives the final JSON, selected schema, route, and source summary, validation report, and unresolved-field list. It checks category--MPN consistency, parameter plausibility, critical missing fields, and weak or conflicting support. The judge may approve, reject, or request review, but it cannot add or alter parameter values. Human review is required when validation fails, the judge flags the case, important fields remain unresolved, sources conflict, or the result depends only on weak fallback evidence. Accepted human corrections are authoritative.

\paragraph{Trace-driven improvement.}
Each item produces a trace containing its route, schema, available evidence, fallback demonstrations when used, output, conflicts, judge verdict, and human decision. The judge is a gate, not a source of truth: it may route an item to review but cannot create labels or change parameter values. Only accepted human decisions can update persistent memory:
\[
\mathcal{M}_{t+1}=\mathcal{M}_t\cup
\{(x_i,y_i^{human},\mathrm{trace}_i): q_i=\mathrm{accepted}\}.
\]
Rejected cases and judge-only approvals are not promoted. Accepted or corrected cases may update the MPN-category store and the validated demonstration memory. GEPA uses accepted and corrected traces, judge errors, and human decisions to revise the judge prompt between evaluation rounds. Therefore, GEPA remains an offline optimization step rather than an additional per-item judging stage. An update is promoted only when it improves held-out judge accuracy or reduces review-triggering failures without increasing false approvals of unsupported outputs.

\section{Experiments}
\label{sec:experiments}

\subsection{Experimental Setup}

Our evaluation asks whether DynaContext improves heterogeneous parameter extraction over a static-prompting pipeline, whether prompt optimization and dynamic retrieval provide complementary gains, why category-specific extraction approaches the accuracy ceiling, and whether dynamic context improves the quality-cost trade-off for smaller inference models.

\paragraph{Benchmarks.}
We use two separately constructed benchmarks (Table~\ref{tab:benchmark-summary}). The heterogeneous benchmark contains 133 production-like component offer records spanning 13 level-3 categories; 131 of them carry at least one schema-supported field, giving 850 non-null gold field-value facts. Gold values were labeled by electronic cost-engineering experts, the same practitioners who consume the extracted parameters downstream. It is the primary test of the paper's title claim: parameter extraction is hard because the active schema changes across capacitors, resistors, inductors, transistors, regulators, filters, and other categories. The ceramic-capacitor benchmark contains 142 records from one component family and isolates the second stage of the method: once the category is known, can DynaContext choose the right local extraction context and approach the accuracy ceiling?

\begin{table}[t]
\centering
\footnotesize
\begin{tabularx}{\columnwidth}{l c X}
\toprule
Benchmark & Records & Evaluation scope \\
\midrule
Ceramic capacitor & 142 & One component family; 7 target parameters per record \\
Multi-component & 133 & 13 level-3 categories; 850 gold facts over 131 scored records \\
\bottomrule
\end{tabularx}
\caption{Benchmark summary for the single-component and heterogeneous
multi-component evaluations.}
\label{tab:benchmark-summary}
\end{table}

\paragraph{Extraction cores and conditions.}
The extraction core is the instruction text occupying the $A^\star_{\hat{c}_i,t}$ slot of Section~\ref{sec:method}. We compare three: the unoptimized Base Prompt, which is the control, and the optimized cores GEPA and SkillOpt (Section~\ref{sec:related-work}).

The single-component sweep runs all three cores, each crossed with four fallback-context settings: no retrieved demonstrations (None), semantic Vector retrieval, lexical BM25 retrieval, and deduplicated Hybrid retrieval. The multi-component sweep runs Base Prompt and SkillOpt, the core deployed in the system of Figure~\ref{fig:upe-flow}, each crossed with demonstrations on or off; this is the two-factor ablation, and its control cell is the base prompt with no demonstrations. GEPA is evaluated only in the controlled single-component study. All nonzero-$k$ runs there use Hybrid retrieval and vary only the number of demonstrations ($k1$, $k3$, $k5$, and $k10$), which isolates a further trade-off: too few demonstrations may not cover the active schema, while too many add distracting fields from neighboring categories.

Retrieved demonstrations come from the validated memory $\mathcal{M}_t$ (Section~\ref{sec:method}) and are human-verified corrections accumulated in operation. The memory is built from a disjoint set of records and contains none of the evaluation data, so no evaluation record can be retrieved as its own demonstration.

\paragraph{The Static-10 reference.}
Static-10 is the pipeline that was running in production before this work. It prepends the same 10 hand-picked demonstrations to every record, irrespective of the component family, and uses neither an optimized core nor retrieval. We report it because it fixes the operating point that the method had to beat.

\paragraph{Metrics.}
The two settings use different data subsets and aggregation levels, so their numbers should not be compared directly. The single-component study reports accuracy, because every record is evaluated against the same seven-parameter ceramic-capacitor schema, making record-level and field-level correctness directly comparable. The multi-component study reports field-level F1, with precision and recall computed over matched predicted field-value facts, because schemas and populated fields vary by category and fact-level scoring captures both missed gold facts and hallucinated fields. We also report estimated inference cost and wall-clock runtime, computed from logged token counts and recorded per-model prices. Confidence intervals over deployments are $t$-intervals on the eight per-model scores; record-level intervals use a cluster bootstrap that resamples component records with replacement over 2000 draws. The main paper reports averages and best cases.

\paragraph{Models and comparability.}
Both benchmarks are run on the same eight deployments: GPT-3.5 Turbo, GPT-4o, GPT-4o mini, GPT-4.1 mini, GPT-5.4, GPT-5.4 mini, GPT-5.5, and DeepSeek-V4-Pro. The model is a sweep factor, not a comparison method. Every condition runs on all eight, so the design is paired: each reported average is over the identical model set, and the difference between two conditions is a method effect measured on the same deployments. We make no cross-model method claims outside Table~\ref{tab:cost-runtime}, which compares deployment choices rather than methods.

All experiments use structured output under the closed-world, schema-constrained contract of Section~\ref{subsec:output-representation}, and both studies fix the gold category and schema so that only the extraction core and the retrieval configuration vary. No human intervenes in a measured run, so the reported scores isolate extraction quality, and human effort enters only through the previously accumulated memory.

\subsection{Single-Component Experiments}
\label{subsec:single-exp}

Each of the three cores is run without retrieved demonstrations and with Vector, BM25, and Hybrid retrieval across the eight deployments. Table~\ref{tab:main-results} reports aggregate and best-case values, which show whether dynamic demonstrations help beyond prompt optimization and whether category-specific extraction approaches the accuracy ceiling.

\begin{table}[t]
\centering
\footnotesize
\begin{tabularx}{\columnwidth}{Xccc}
\toprule
Core, retrieval & Mean acc. & [95\% CI] & Best run \\
\midrule
Base Prompt, none & 86.64 & [82.6, 90.7] & 90.14 \\
Base Prompt, best & 94.83 & [92.5, 97.1] & 98.79 \\
GEPA, none & 95.56 & [95.0, 96.1] & 96.68 \\
GEPA, best & 98.49 & [98.0, 99.0] & \textbf{99.50} \\
SkillOpt, none & 96.93 & [96.4, 97.4] & 97.89 \\
SkillOpt, best & \textbf{98.59} & [98.3, 98.9] & 99.20 \\
\bottomrule
\end{tabularx}
\caption{Single-component ceramic-capacitor accuracy (\%). ``None'' is the
core with no retrieved demonstrations; ``best'' is its strongest retrieval
policy, Hybrid except for GEPA, where Vector wins. Mean and interval are over
the eight deployments; ``Best run'' is the single strongest deployment for
that condition.}
\label{tab:main-results}
\end{table}

Across the sweep, retrieved context raises average accuracy from 93.05\% in the no-retrieval conditions to 97.09\%. The strongest single result combines DynaContext with GEPA, reaching 99.50\% with DeepSeek-V4-Pro and Vector retrieval. The smaller GPT-5.4 mini model reaches 99.30\% with GEPA and BM25 retrieval and 99.20\% with SkillOpt and Hybrid retrieval, showing that an optimized core with local demonstrations can substitute for larger-model capacity when the active schema is fixed.

\subsection{Multi-Component Experiments}
\label{subsec:multi-exp}

Table~\ref{tab:cross-component-results} evaluates heterogeneous description-only parameter extraction over the 850 gold field-value facts, crossing the Base Prompt and SkillOpt cores with demonstrations on or off across the eight deployments. Both factors help on their own. Against the demonstration-free base prompt at 51.76 F1, dynamic demonstrations alone add 7.47 points [6.42, 8.51] and the optimized core alone adds 15.14 [10.66, 19.61], while the two together gain 19.22 [14.18, 24.27]; every effect is positive in all eight deployments (Wilcoxon signed-rank, $p=0.0039$). The joint effect is smaller than the sum of the parts: demonstrations are worth 7.47 points on the base prompt but only 4.09 [2.18, 5.99] on top of the optimized core, so the two mechanisms supply partly overlapping information about schema, formatting, and null behavior rather than independent gains. These paired intervals, not the marginal ones in Table~\ref{tab:cross-component-results}, are the relevant test: deployments differ more from one another than conditions do within a deployment, so the marginal intervals overlap while every paired effect is significant. Dynamic demonstrations also beat static ones on equal terms, since the Static-10 reference reaches 53.96 with ten fixed demonstrations.

\begin{table}[tb]
\centering
\footnotesize
\begin{tabularx}{\columnwidth}{Xccc}
\toprule
Condition & Mean F1 & [95\% CI] & Best run \\
\midrule
Base prompt, none & 51.76 & [48.9, 54.7] & 55.81 \\
Base prompt, $k5$ & 59.22 & [57.1, 61.4] & 63.33 \\
SkillOpt, none & 66.89 & [61.7, 72.1] & 73.86 \\
SkillOpt, $k5$ & \textbf{70.98} & [65.9, 76.1] & \textbf{77.71} \\
\midrule
Static-10 (reference) & 53.96 & [49.9, 58.0] & 61.66 \\
\bottomrule
\end{tabularx}
\caption{Two-factor ablation on the heterogeneous benchmark: F1 (\%) over 850 gold facts, crossing the extraction core with dynamic demonstrations at fixed $k=5$. Intervals are 95\% $t$-intervals over the eight deployments; paired effects are reported in the text. Static-10 uses a different prompt from the four crossed conditions, so it is shown for reference and is not part of the ablation.}
\label{tab:cross-component-results}
\end{table}

\begin{table}[t]
\centering
\footnotesize
\begin{tabularx}{\columnwidth}{Xccc}
\toprule
Configuration & F1 & Cost (\$) & Time (s) \\
\midrule
GPT-5.5, $k5$ & 75.71 & 2.955 & 768.9 \\
DeepSeek-V4-Pro, $k3$ & \textbf{77.93} & 0.498 & 422.3 \\
GPT-5.4 mini, $k5$ & 77.42 & \textbf{0.203} & \textbf{174.5} \\
\bottomrule
\end{tabularx}
\caption{Cost and runtime of the three strongest configurations. The model
varies across configurations, so this compares deployments rather than methods.}
\label{tab:cost-runtime}
\end{table}

We do not compare Vector and BM25 separately here because either signal alone is incomplete: lexical overlap matters for MPN fragments, packages, and units, semantic similarity for equivalent phrasings. Varying $k$ instead measures how much Hybrid context is useful, and moderate context sizes usually win: $k3$ or $k5$ is optimal for six of the eight models. The strongest single result is DeepSeek-V4-Pro with the SkillOpt core and three Hybrid-retrieved demonstrations, reaching 77.93 F1.

Holding the model fixed, DynaContext costs more than the Static-10 reference in every deployment, between 1.05 and 1.93 times as much and 1.48 times on average, in exchange for a mean gain of 17.32 F1 points. The extra tokens buy accuracy rather than saving money.

Table~\ref{tab:cost-runtime} shows that spending the budget on a larger model is the worst alternative. GPT-5.5 costs roughly fifteen times as much per run as GPT-5.4 mini and still scores 1.72 F1 points lower, and the best result overall, 77.93 with DeepSeek-V4-Pro, costs 2.45 times as much as GPT-5.4 mini for 0.51 points more. A small model with the right schema and demonstrations is thus a better use of the same budget than a large model left to recover that context implicitly.

\subsection{Analysis}
The gap between the two settings is a gap in what the demonstrations have to do. Under a fixed schema, they only supply local formatting guidance, and being drawn from the same family, they reinforce the target schema. Heterogeneously, they must also help resolve the active family, and a similar-looking demonstration from a neighboring category can suggest the wrong fields or unit conventions, which is why dynamic context improves heterogeneous F1 without reaching the fixed-schema ceiling.

We tag failures at the field level so that each error points to the part of the system that should improve. Category, schema, and demonstration-mismatch errors inform the extractor and retrieval policy; MPN, value, unit, datasheet, and external-API errors point to evidence acquisition or normalization failures; judge false approvals and false rejections tune the GEPA-optimized judge; and review-correction labels mark cases where human feedback should be promoted into validated memory.

\FloatBarrier

\section{Limitations}
\label{sec:limitations}
Our benchmarks are small: 142 single-category records and 133 heterogeneous records evaluated over 850 gold facts, drawn from one industrial corpus, so corpus-specific effects cannot be ruled out even where the reported intervals are narrow. The heterogeneous set is also unbalanced, as it mirrors the production mix rather than a designed sample: ceramic capacitors and chip resistors supply 74\% of the gold facts, while five of the thirteen categories contribute fewer than ten each, so per-category estimates for rare families are noisy, and the aggregate is dominated by two families. Evaluation on public benchmarks such as WDC-PAVE \citep{brinkmann2024wdcpave} is left for future work. Several components rely on heuristics with fixed settings, including Jaro--Winkler taxonomy matching and the static interleave used for hybrid demonstration retrieval; threshold sensitivity analyses and per-query retrieval weighting \citep{hsu2025dat} are natural extensions. The system depends on proprietary interfaces, an external parts API, and datasheet retrieval, whose responses change over time; our experiments use cached responses with recorded query dates, and datasheet extraction remains least reliable for family tables with many close variants. The GEPA-optimized judge is audited against human review decisions in operation, but its calibration can drift as the input distribution shifts, so conservative review routing remains necessary for rare categories. The demonstration memory's contribution is measured directly, but improvement of the memory across successive review rounds is not evaluated.

Finally, the underlying offer records cannot be released because of confidentiality constraints.

\section{Conclusion}
\label{sec:conclusion}
DynaContext treats unstructured electronic component parameter extraction as a schema-grounded, source-aware, self-improving agent problem: an offline optimized core is composed at inference time with the resolved schema, gathered evidence, unresolved fields, and validated demonstrations, and every output and every memory update passes deterministic validation, a GEPA-optimized judge, and human review. A two-factor ablation attributes the heterogeneous gain to demonstrations and the optimized core in roughly a one-to-two ratio, and the resulting configuration gains 17.32 F1 over the deployed pipeline across eight deployments. The remaining gap to the fixed-schema ceiling is a schema-resolution gap, which is where we expect further progress.

\bibliography{paper}

\end{document}